\documentclass[conference,a4paper]{IEEEtran}

\usepackage{amsfonts}
\usepackage{cite}
\usepackage{graphicx}
\usepackage{amssymb}
\usepackage{subfigure}
\usepackage{array}
\usepackage{color}
\usepackage{amsmath}
\usepackage{algorithmic}
\usepackage{algorithm}
\usepackage{cases}
\usepackage{mathrsfs}
\usepackage{epsfig}
\usepackage{psfrag}
\usepackage{url}
\usepackage{setspace}
\usepackage{stfloats}
\usepackage{enumerate}
\usepackage{enumitem}
\allowdisplaybreaks[4]
\usepackage{multirow}

\usepackage{etoolbox}
\makeatletter
\patchcmd{\@makecaption}
  {\scshape}
  {}
  {}
  {}
\makeatletter
\patchcmd{\@makecaption}
  {\\}
  {.\ }
  {}
  {}
\makeatother

\newtheorem{Thm}{Theorem}
\newtheorem{Lem}{Lemma}

\newtheorem{Prob}{Problem}
\newtheorem{Rem}{Remark}

\newtheorem{Asump}{Assumption}

\IEEEoverridecommandlockouts

\begin{document}

\title{Sample-based Federated Learning via Mini-batch SSCA}
\author{\IEEEauthorblockN{Chencheng Ye} \IEEEauthorblockA{Shanghai Jiao Tong University, China}
\and \IEEEauthorblockN{Ying Cui} \IEEEauthorblockA{Shanghai Jiao Tong University, China}\thanks{This work was supported in part by the Natural Science Foundation of Shanghai under Grant 20ZR1425300. (Corresponding author: Ying Cui.)}
}


\maketitle

\begin{abstract}
	In this paper, we investigate unconstrained and constrained sample-based federated optimization, respectively. For each problem, we propose a privacy preserving algorithm using stochastic successive convex approximation (SSCA) techniques, and show that it can converge to a Karush-Kuhn-Tucker (KKT) point. To the best of our knowledge, SSCA has not been used for solving federated optimization, and federated optimization with nonconvex constraints has not been investigated. Next, we customize the two proposed SSCA-based algorithms to two application examples, and provide closed-form solutions for the respective approximate convex problems at each iteration of SSCA. Finally, numerical experiments demonstrate inherent advantages of the proposed algorithms in terms of convergence speed, communication cost and model specification.
\end{abstract}


\section{Introduction}
\label{sec:intro}
Machine learning with distributed databases has been a hot research area~\cite{li2014communication}.
The amount of data at each client can be large, and hence the data uploading to a central server may be constrained by energy and bandwidth limitations.
Besides, local data may contain highly sensitive information, e.g., travel records, health information and web browsing history, and thus a client may be unwilling to share it. Therefore, it is impossible or undesirable to upload distributed databases to a central server.
Recent years have witnessed the growing interest in federated learning, where data is maintained locally during the collaborative training of the server and clients~\cite{yang2019federated}. Data privacy and communication efficiency are the two main advantages of federated learning, as only model parameters or gradients are exchanged in the training process. 

Most existing works for federated learning focus on solving unconstrained optimization problems using mini-batch stochastic gradient descent (SGD)~\cite{mcmahan2017communication,yang2019scheduling,yu2019parallel,yang2019federated,hardy2017private}. Depending on whether data is distributed over the sample space or feature space, federated learning can be typically classified into sample-based (horizontal) federated learning and feature-based (vertical) federated learning~\cite{yang2019federated}. 
In sample-based federated learning, the datasets of different clients have the same feature space but little intersection on the sample space. Most studies on federated learning focus on this category~\cite{mcmahan2017communication,yang2019scheduling,yu2019parallel}. In the existing sample-based federated learning algorithms, the global model is iteratively updated at the server by aggregating and averaging the locally computed models at clients. Data privacy is naturally preserved as the model averaging steps avoid exposing raw data.
Specifically, at each communication round, the selected clients download the current model parameters and conduct one or multiple SGD updates to refine the local model. Multiple local SGD updates can reduce the required number of model averaging steps and hence save communication costs. However, they may yield the divergence of sample-based federated learning when local datasets across clients are heterogeneous. The most commonly used sample-based federated learning algorithm is the Federated Averaging algorithm~\cite{mcmahan2017communication}. 
On the contrary, in feature-based federated learning, the datasets of different clients share the same sample space but differ in the feature space. Feature-based federated learning is more challenging, as a client cannot obtain the gradient of a loss function relying purely on its local data. In the existing feature-based federated learning algorithms~\cite{yang2019federated,hardy2017private}, intermediate parameters are exchanged for calculating the gradient before model aggregation steps.

SGD has long been used for solving unconstrained stochastic optimization problems or stochastic optimization problems with deterministic convex constraints. Recently, stochastic successive convex approximation (SSCA) is proposed to obtain Karush-Kuhn-Tucker (KKT) points of stochastic optimization problems with deterministic convex constraints~\cite{Yang} and with general stochastic nonconvex constraints~\cite{Liu,Ye}. Apparently, SSCA has a wider range of applications than SGD. It has also been shown in\cite{Yang} that SSCA empirically achieves a higher convergence speed than SGD, as SGD utilizes only first-order information of the objective function.
Some recent works have applied SSCA to solve machine learning problems~\cite{scardapane2018stochastic}. Nevertheless, SSCA has not been applied for solving federated optimization so far.

In this paper, we focus on designing sample-based federated learning algorithms using SSCA for unconstrained problems and constrained problems, respectively. 
First, we propose a privacy preserving algorithm to obtain a KKT point of unconstrained sample-based federated optimization using mini-batch SSCA, and analyze its computational complexity and convergence. Such algorithm empirically converges faster (i.e., achieves a lower communication cost)  than the SGD-based ones in~\cite{mcmahan2017communication,yang2019scheduling,yu2019parallel} and can achieve the same order of computational complexity as the SGD-based ones in~\cite{mcmahan2017communication,yang2019scheduling,yu2019parallel}. 
Then, we propose a privacy preserving algorithm to obtain a KKT point of constrained sample-based federated optimization by combining the exact penalty method for SSCA in~\cite{Ye} and mini-batch techniques, and analyze its convergence.
Notice that federated optimization with nonconvex constraints, which can explicitly limit the cost function of a model, has not been investigated so far. 
Next, we customize the two SSCA-based algorithms to two application examples, and show that all updates at each iteration have closed-form expressions.
Finally, numerical experiments demonstrate that the proposed algorithm for unconstrained sample-based federated optimization converges faster (i.e., yield lower communication costs) than the existing SGD-based ones~\cite{mcmahan2017communication,yang2019scheduling,yu2019parallel}, and the proposed algorithms for constrained federated optimization can more flexibly specify a training model.

\section{System Setting}
Consider $N$ data samples, each of which has $K$ features. For all $n\in\mathcal N\triangleq\{1,\dots,N\}$, the $K$ features of the $n$-th sample are represented by a $K$-dimensional vector $\mathbf{x}_n\in\mathbb{R}^{K}$.
Consider a central server connected with $I$ local clients, each of which maintains a local dataset.
Specifically, partition $\mathcal{N}$ into $K$ disjoint subsets, denoted by $\mathcal{N}_i$, $i\in\mathcal{I}\triangleq\{1,\dots,I\}$, where $N_i\triangleq|\mathcal{N}_i|$ denotes the cardinality of the $i$-th subset and $\sum_{i\in\mathcal{I}}N_i=N$.
For all $i\in\mathcal{I}$, the $i$-th client maintains a local dataset containing $N_i$ samples, i.e., $\mathbf{x}_n$, $n\in\mathcal{N}_i$.
For example, two companies with similar business in different cities may have different user groups (from their respective regions) but the same type of data, e.g., users' occupations, ages, incomes, deposits, etc.
The server and $I$ clients collaboratively train a model from the local datasets stored on the $I$ clients under the condition that each client cannot expose its local raw data to the others.
This training process is referred to as sample-based (horizontal) federated learning~\cite{yang2019federated}.
The underlying optimization, termed sample-based federated optimization~\cite{yang2019federated}, is to minimize the following function:
\begin{align}
	&F_{0}(\boldsymbol\omega)\triangleq
	\frac{1}{N}\sum_{n\in\mathcal{N}}
	f_{0}(\boldsymbol\omega,\mathbf{x}_n)\label{eqn:Fs0}
\end{align}
with respect to (w.r.t.) model parameters $\boldsymbol\omega\in\mathbb{R}^d$.
To be general, we do not assume $F_{0}(\boldsymbol\omega)$ to be convex in $\boldsymbol\omega$.

In Section~\ref{sec:uncon} and Section~\ref{sec:con}, we investigate sample-based federated Learning for unconstrained optimization and constrained optimizaiton, respectively.
To guarantee the convergence of the proposed SSCA-based federated learning algorithms, we assume that $f_{0}\left(\boldsymbol\omega,\mathbf{x}_n\right)$ satisfies the following assumption in the rest of the paper.
\begin{Asump}[Assumption on $f(\boldsymbol\omega,\mathbf{x})$]\label{asump:f}\cite{Yang,Liu}
	For any given $\mathbf{x}$, each $f(\boldsymbol\omega,\mathbf{x})$ is continuously differentiable, and its gradient is Lipschitz continuous.	
\end{Asump}
\begin{Rem}[Discussion on Assumption~\ref{asump:f}]
	Assumption~\ref{asump:f} is also necessary for the convergence of SSCA~\cite{Yang,Liu,Ye} and SGD~\cite{yu2019parallel}.
\end{Rem}

\section{Sample-based Federated Learning for Unconstrained Optimization}\label{sec:uncon}
In this section, we consider the following unconstrained sample-based federated optimization problem:
\begin{Prob}[Unconstrained Sample-based Federated Optimization]\label{Prob:uncon-sample}
	\begin{align}
		&\min_{\boldsymbol\omega} F_{0}(\boldsymbol\omega)\nonumber
	\end{align}
	where $F_{0}(\boldsymbol\omega)$ is given by~\eqref{eqn:Fs0}.
\end{Prob}

Problem~\ref{Prob:uncon-sample} (whose objective function has a large number of terms) is usually transformed to an equivalent stochastic optimization problem, and solved using stochastic optimization algorithms.
The SGD-based algorithms in~\cite{mcmahan2017communication,yang2019scheduling,yu2019parallel}, proposed to obtain a KKT point of Problem~\ref{Prob:uncon-sample}, may have unsatisfactory convergence speeds and high communication costs, as SGD only utilizes the first-order information of an objective function.
In the following, we propose a privacy-preserving sample-based federated learning algorithm, i.e., Algorithm~\ref{alg:uncon-sample}, to obtain a KKT point of Problem~\ref{Prob:uncon-sample} using mini-batch SSCA. 
It has been shown in~\cite{Yang} that SSCA empirically achieves a higher convergence speed than SGD. Later in Section~\ref{sec:simu}, we shall numerically show that the proposed SSCA-based algorithm converges faster than the SGD-based algorithms in~\cite{mcmahan2017communication,yang2019scheduling,yu2019parallel}.

\begin{algorithm}[t]
	\caption{Mini-batch SSCA for Problem~\ref{Prob:uncon-sample}}
	\begin{algorithmic}[1]
		\STATE \textbf{initialize}: choose any ${\boldsymbol\omega}^{1}$ at the server.\\
		\FOR{$t=1,2,\dots,T-1$}
		\STATE the server sends ${\boldsymbol\omega}^{(t)}$ to all clients.
		\STATE for all $i\in\mathcal{I}$, client $i$ randomly selects a mini-batch $\mathcal{N}^{(t)}_i\subseteq\mathcal{N}_i$, computes $\mathbf q_{0}\left({\boldsymbol\omega}^{(t)},(\mathbf{x}_n)_{n\in\mathcal{N}^{(t)}_i}\right)$ and sends it to the server.
		\STATE the server obtains $\bar{\boldsymbol\omega}^{(t)}$ by solving Problem~\ref{Prob:uncon-sample-ap}, and updates ${\boldsymbol\omega}^{(t+1)}$ according to \eqref{eqn:updatew}.
		\ENDFOR
		\STATE \textbf{Output}: ${\boldsymbol\omega}^T$
	\end{algorithmic}\label{alg:uncon-sample}
\end{algorithm}

\subsection{Algorithm Description}
The main idea of Algorithm~\ref{alg:uncon-sample} is to solve a sequence of successively refined convex problems, each of which is obtained by approximating $F_{0}(\boldsymbol\omega)$ with a convex function based on its structure and samples in a randomly selected mini-batch by the server.
Specifically, at iteration $t$, we choose: 
\begin{align}
	\bar F^{(t)}_{0}(\boldsymbol\omega)=&(1-\rho^{(t)})\bar{F}^{(t-1)}_{0}(\boldsymbol\omega)\nonumber\\
	&+\rho^{(t)}\sum_{i\in\mathcal{I}}\frac{N_i}{BN}\sum_{n\in\mathcal N_i^{(t)}}\bar{f}_{0}(\boldsymbol\omega,{\boldsymbol\omega}^{(t)},\mathbf{x}_n)\label{eqn:Fs0bar}
\end{align}
with $\bar F_{0}^{(0)}(\boldsymbol\omega)=0$ as an approximation function of $F_{0}(\boldsymbol\omega)$,
where $\rho^{(t)}$ is a stepsize satisfying:
\begin{align}
	&\rho^{(t)}>0,\quad \lim_{t\to\infty}\rho^{(t)}=0,\quad \sum_{t=1}^\infty\rho^{(t)}=\infty,\label{eqn:rho}
\end{align}
$\mathcal N_i^{(t)}\subseteq\mathcal N_i$ is a randomly selected mini-batch by client $i$ at iteration $t$, $B\leq N_i$ is the batch size, and $\bar{f}_{0}(\boldsymbol\omega,{\boldsymbol\omega}^{(t)},\mathbf{x}_n)$ is a convex approximation of $f_{0}(\boldsymbol\omega,\mathbf{x}_n)$ around ${\boldsymbol\omega}^{(t)}$ satisfying the following assumptions. A common example of $\bar{f}_{0}$ will be given later.
\begin{Asump}[Assumptions on $\bar{f}(\boldsymbol\omega,\boldsymbol\omega',\mathbf{x})$ for Approximating $f(\boldsymbol\omega,\mathbf{x})$ Around $\boldsymbol\omega'$]\label{asump:fbar}~\cite{Liu}
	1) $\nabla \bar{f}(\boldsymbol\omega,\boldsymbol\omega,\mathbf{x})=\nabla f(\boldsymbol\omega,\mathbf{x})$;
	2) $\bar{f}(\boldsymbol\omega,\boldsymbol\omega',\mathbf{x})$ is strongly convex in $\boldsymbol\omega$;
	3) $\bar{f}(\boldsymbol\omega,\boldsymbol\omega',\mathbf{x})$ is Lipschitz continuous in both $\boldsymbol\omega$ and $\boldsymbol\omega'$;
	4) $\bar{f}(\boldsymbol\omega,\boldsymbol\omega',\mathbf{x})$, its derivative, and its second order derivative w.r.t. $\boldsymbol\omega$ are uniformly bounded.
\end{Asump}

Assumption~\ref{asump:fbar} is necessary for the convergence of SSCA~\cite{Yang,Liu}.
Note that for all $i\in\mathcal{I}$ and any mini-batch $\mathcal{N}'_i\subseteq\mathcal{N}_i$ with batch size $B\leq N_i$, $\sum_{n\in\mathcal{N}'_i}\bar{f}_{0}(\boldsymbol\omega,\boldsymbol\omega',\mathbf{x}_n)$ can be written as $\sum_{n\in\mathcal{N}'_i}\bar{f}_{0}(\boldsymbol\omega,\boldsymbol\omega',\mathbf{x}_n)
=p_{0}\left(\mathbf q_{0}\left(\boldsymbol\omega',(\mathbf{x}_n)_{n\in\mathcal{N}'_i}\right),\boldsymbol\omega\right)$ with $p_{0}:\mathbb{R}^{D_0+d}\to\mathbb{R}$ and $\mathbf q_{0}:\mathbb{R}^{BK+d}\to\mathbb{R}^{D_0}$.
Assume that the expressions of $\bar{f}_{0}$, $p_{0}$ and $\mathbf q_{0}$ are known to the server and $N$ clients. Each client $i\in\mathcal{I}$ computes $\mathbf q_{0}\left({\boldsymbol\omega}^{(t)},(\mathbf{x}_n)_{n\in\mathcal{N}^{(t)}_i}\right)$ and sends it to the server. Then, the server solves the following convex approximate problem to obtain $\bar{\boldsymbol\omega}^{(t)}$.
\begin{Prob}[Convex Approximate Problem of Problem~\ref{Prob:uncon-sample}]\label{Prob:uncon-sample-ap}
	\begin{align}
		&\bar{\boldsymbol\omega}^{(t)}\triangleq\mathop{\arg\min}_{\boldsymbol\omega} \bar F_{0}^{(t)}(\boldsymbol\omega)\nonumber
	\end{align}
\end{Prob}

Problem~\ref{Prob:uncon-sample-ap} is convex and can be solved with conventional convex optimization techniques.
Given $\bar{\boldsymbol\omega}^{(t)}$, the server updates ${\boldsymbol\omega}^{(t)}$ according to:
\begin{align}
	&{\boldsymbol\omega}^{(t+1)}=(1-\gamma^{(t)}){\boldsymbol\omega}^{(t)}+\gamma^{(t)}\bar{\boldsymbol\omega}^{(t)},\ t=1,2,\dots \label{eqn:updatew}
\end{align}
where $\gamma^{(t)}$ is a stepsize satisfying:
\begin{align}
	&\gamma^{(t)}=0,\ \lim_{t\to\infty}\gamma^{(t)}=0,\  \sum_{t=1}^\infty\gamma^{(t)}=\infty,\nonumber\\
	&\sum_{t=1}^\infty\left(\gamma^{(t)}\right)^2<\infty,\quad \lim_{t\to\infty}\frac{\gamma^{(t)}}{\rho^{(t)}}=0.\label{eqn:gamma}
\end{align}

The detailed procedure is summarized in Algorithm~\ref{alg:uncon-sample}, and the convergence of Algorithm~\ref{alg:uncon-sample} is summarized below.
\begin{Thm}[Convergence of Algorithm~\ref{alg:uncon-sample}]\label{thm:uncon-sample}
	Suppose that $f_{0}$ satisfies Assumption~\ref{asump:f}, $\bar{f}_{0}$ satisfies Assumption~\ref{asump:fbar}, and the sequence $\{{\boldsymbol\omega}^{(t)}\}$ generated by Algorithm~\ref{alg:uncon-sample} is bounded. Then, every limit point of $\{{\boldsymbol\omega}^{(t)}\}$ is a KKT point of Problem~\ref{Prob:uncon-sample} almost surely.
\end{Thm}
\begin{IEEEproof}[Proof (Sketch)]
	It follows from~\cite[Lemma1]{Lemma} that $\lim_{t\to\infty}\Vert\nabla\bar{F}_0^t(\boldsymbol\omega^t)$
	$-\nabla F_0(\boldsymbol\omega^t)\Vert=0$. Then, the convergence of Algorithm~\ref{alg:uncon-sample} can be obtained by generalizing the analysis in \cite[Theorem~1]{Yang}. 
\end{IEEEproof}

\subsection{Security Analysis}
We establish the security of Algorithm~\ref{alg:uncon-sample}.
If for all $i\in\mathcal{I}$ and any mini-batch $\mathcal{N}'_i\subseteq\mathcal{N}_i$ with batch size $B\leq N_i$, the system of equations w.r.t. $\mathbf{z}\in\mathbb{R}^{BK}$, i.e., $\mathbf q_{0}\left(\boldsymbol\omega',\mathbf{z}\right)=\mathbf q_{0}\left(\boldsymbol\omega',(\mathbf{x}_n)_{n\in\mathcal{N}'_i}\right)$, has an infinite (or a sufficiently large) number of solutions, then raw data $\mathbf{x}_n$, $n\in\mathcal{N}^{(t)}_i$ cannot be extracted from $\mathbf q_{0}\left({\boldsymbol\omega}^{(t)},(\mathbf{x}_n)_{n\in\mathcal{N}^{(t)}_i}\right)$ in Step 4 of Algorithm~\ref{alg:uncon-sample}, and hence Algorithm~\ref{alg:uncon-sample} can preserve data privacy.
Otherwise, extra privacy mechanisms, such as homomorphic encryption and secret sharing, can be applied to preserve data privacy.

\subsection{Algorithm Example}
Finally, we provide an example of $\bar{f}_{0}$ which satisfies Assumption~\ref{asump:fbar} and yields an analytical solution of Problem~\ref{Prob:uncon-sample}:
\begin{align}
	\bar{f}_{0}(\boldsymbol\omega,{\boldsymbol\omega}^{(t)}\!,\mathbf{x}_n)\!=\!&\left(\nabla f_{0}({\boldsymbol\omega}^{(t)}\!,\mathbf{x}_n)\right)^T\!\!\left(\boldsymbol\omega\!-\!{\boldsymbol\omega}^{(t)}\!\right)\!+\!\tau\Vert{\boldsymbol\omega\!-\!{\boldsymbol\omega}^{(t)}}\!\Vert_2^2, \label{eqn:fs0bar}
\end{align}
where $\tau>0$ can be any constant, and the term $\tau\Vert{\boldsymbol\omega-{\boldsymbol\omega}^{(t)}}\Vert_2^2$ is used to ensure strong convexity.
Obviously, $\bar{f}_{0}$ given by~\eqref{eqn:fs0bar} satisfies Assumption~\ref{asump:fbar}.
Notice that Problem~\ref{Prob:uncon-sample-ap} with $\bar{f}_{0}$ given by~\eqref{eqn:fs0bar} is an unconstrained convex quadratic programming w.r.t. $\boldsymbol\omega$ and hence has an analytical solution with the same order of computational complexity as the SGD-based ones in~\cite{mcmahan2017communication,yang2019scheduling,yu2019parallel} (which is $\mathcal O(d)$). The details of the analytical solution will be given in Section~\ref{sec:application}.

\section{Sample-based Federated Learning for Constrained Optimization}\label{sec:con}
In this section, we consider the following constrained sample-based federated optimization problem:
\begin{Prob}[Constrained Sample-based Federated Optimization]\label{Prob:con-sample}
	\begin{align}
		\min_{\boldsymbol\omega}\ &F_{0}(\boldsymbol\omega)\nonumber\\
		\text{s.t.}\ &F_{m}(\boldsymbol\omega)\leq 0,\quad m=1,2,\dots,M,\nonumber
	\end{align}
	where $F_{0}(\boldsymbol\omega)$ is given by~\eqref{eqn:Fs0}, 
	and
	\begin{align}
		F_{m}(\boldsymbol\omega)\triangleq\frac{1}{N}\sum_{n\in\mathcal{N}} f_{m}(\boldsymbol\omega,\mathbf{x}_n),\quad m=1,2,\dots,M.\nonumber
	\end{align}
\end{Prob}

To be general, $F_{m}(\boldsymbol\omega)$, $m=0,\dots,M$ are not assumed to be convex in $\boldsymbol\omega$. Notice that federated optimization with nonconvex constraints has not been investigated so far. It is quite challenging, as the stochastic nature of a constraint function may cause infeasibility at each iteration of an ordinary stochastic iterative method~\cite{Ye}.
In the following, we propose a privacy-preserving sample-based federated learning algorithm, i.e., Algorithm~\ref{alg:con-sample}, to obtain a KKT point of Problem~\ref{Prob:con-sample}, by combining the exact penalty method~\cite{bertsekas1998nonlinear} for SSCA in~\cite{Ye} and mini-batch techniques.

\subsection{Algorithm Description}
\begin{algorithm}[t]
	\caption{Mini-batch SSCA for Problem~\ref{Prob:con-sample}}
	\begin{algorithmic}[1]
		\STATE \textbf{initialize}: choose any ${\boldsymbol\omega}^{1}$ and $c>0$ at the server.\\
		\FOR{$t=1,2,\dots,T-1$}
		\STATE the server sends ${\boldsymbol\omega}^{(t)}$ to all clients.
		\STATE for all $i\in\mathcal{I}$, client $i$ randomly selects a mini-batch $\mathcal{N}^{(t)}_i\subseteq\mathcal{N}_i$, computes $\mathbf q_{m}\left({\boldsymbol\omega}^{(t)},(\mathbf{x}_n)_{n\in\mathcal{N}^{(t)}_i}\right)$, $m=0,1,\dots,M$ and sends them to the server.
		\STATE the server obtains $(\bar{\boldsymbol\omega}^{(t)}, \mathbf s^{(t)})$ by solving Problem~\ref{Prob:con-sample-ap}, and updates ${\boldsymbol\omega}^{(t+1)}$ according to \eqref{eqn:updatew}.
		\ENDFOR
		\STATE \textbf{Output}: ${\boldsymbol\omega}^T$
	\end{algorithmic}\label{alg:con-sample}
\end{algorithm}

First, we transform Problem~\ref{Prob:con-sample} to the following stochastic optimization problem whose objective function is the weighted sum of the original objective and the penalty for violating the original constraints.
\begin{Prob}[Transformed Problem of Problem~\ref{Prob:con-sample}]\label{Prob:con-sample-ep}
	\begin{align}
		\min_{\boldsymbol\omega,\mathbf s}\quad &F_{0}(\boldsymbol\omega)+c\sum_{m=1}^M s_m\nonumber\\
		\text{s.t.}\quad &F_{m}(\boldsymbol\omega)\leq s_m,\quad m=1,2,\dots,M,\nonumber\\
		&s_m\geq 0,\quad m=1,2,\dots,M,\nonumber
	\end{align}
	where $\mathbf{s}\triangleq(s_m)_{m=1,\dots,M}$ are slack variables and $c>0$ is a penalty parameter that trades off the original objective function and the slack penalty term.
\end{Prob}

At iteration $t$, we choose $\bar F^{(t)}_{0}(\boldsymbol\omega)$ given in~\eqref{eqn:Fs0bar} as an approximation function of $F_{0}(\boldsymbol\omega)$, and choose:
\begin{align}
	\bar F^{(t)}_{m}(\boldsymbol\omega)=&(1-\rho^{(t)})\bar{F}^{(t-1)}_{m}(\boldsymbol\omega)+\rho^{(t)}
	\sum_{i\in\mathcal{I}}\frac{N_i}{BN}\nonumber\\
	&\times\sum_{n\in\mathcal N_i^{(t)}}\bar{f}_{m}(\boldsymbol\omega,{\boldsymbol\omega}^{(t)},\mathbf{x}_n),\quad m=1,\dots,M\label{eqn:Fsmbar}
\end{align}
with $\bar F_{m}^{(0)}(\boldsymbol\omega)=0$ as an approximation function of $F_{m}(\boldsymbol\omega)$, for all $m=1,\dots,M$, where $\rho^{(t)}$ is a stepsize satisfying~\eqref{eqn:rho},
$\mathcal N^{(t)}_i$ is the randomly selected mini-batch by client $i$ at iteration $t$,
and $\bar{f}_{m}(\boldsymbol\omega,{\boldsymbol\omega}^{(t)},\mathbf{x}_n)$ is a convex approximation of $f_{m}(\boldsymbol\omega,\mathbf{x}_n)$ around ${\boldsymbol\omega}^{(t)}$ satisfying $\bar{f}_{m}(\boldsymbol\omega,\boldsymbol\omega,\mathbf{x})=f_{m}(\boldsymbol\omega,\mathbf{x})$ and Assumption~\ref{asump:fbar} for all $m=1,\dots,M$.
A common example of $\bar{f}_{m}$, $m=0,\dots,M$ will be given later.

Note that for all $i\in\mathcal{I}$ and any mini-batch $\mathcal{N}'_i\subseteq\mathcal{N}_i$ with batch size $B\leq N_i$, $\sum_{n\in\mathcal{N}'_i}\bar{f}_{m}(\boldsymbol\omega,\boldsymbol\omega',\mathbf{x}_n)$, $m=0,\dots,M$ can be written as $\sum_{n\in\mathcal{N}'_i}\bar{f}_{m}(\boldsymbol\omega,\boldsymbol\omega',\mathbf{x}_n)
=p_{m}\left(\mathbf q_{m}\left(\boldsymbol\omega',(\mathbf{x}_n)_{n\in\mathcal{N}'_i}\right),\boldsymbol\omega\right)$, $m=0,\dots,M$, with $p_{m}:\mathbb{R}^{D_m+d}\to\mathbb{R}$ and $\mathbf q_{m}:\mathbb{R}^{BK+d}\to\mathbb{R}^{D_m}$.
Assume that the expressions of $\bar{f}_{m}$, $p_{m}$ and $\mathbf q_{m}$, $m=0,\dots,M$ are known to the server and $N$ clients.  Each client $i\in\mathcal{I}$ computes $\mathbf q_{m}\left({\boldsymbol\omega}^{(t)},(\mathbf{x}_n)_{n\in\mathcal{N}^{(t)}_i}\right)$, $m=0,\dots,M$ and send them to the server. Then, the server solves the following approximate problem to obtain $\bar{\boldsymbol\omega}^{(t)}$.

\begin{Prob}[Convex Approximate Problem of Problem~\ref{Prob:con-sample-ep}]\label{Prob:con-sample-ap}
	\begin{align}
		(\bar{\boldsymbol\omega}^{(t)},\mathbf{s}^{(t)})\triangleq&\mathop{\arg\min}_{\boldsymbol\omega,\mathbf{s}} \bar F^{(t)}_{0}(\boldsymbol\omega)+c\sum_{m=1}^M s_m\nonumber\\
		\text{s.t.}\quad &\bar F^{(t)}_{m}(\boldsymbol\omega)\leq s_m,\quad m=1,2,\dots,M,\nonumber\\
		&s_m\geq 0,\quad m=1,2,\dots,M.\nonumber
	\end{align}
\end{Prob}

Problem~\ref{Prob:con-sample-ap} is convex and can be readily solved.
Given $\bar{\boldsymbol\omega}^{(t)}$, the server updates ${\boldsymbol\omega}^{(t)}$ according to \eqref{eqn:updatew}. 
The detailed procedure is summarized in Algorithm~\ref{alg:con-sample}, and the convergence of Algorithm~\ref{alg:con-sample} is summarized below. Consider a sequence $\{c_j\}$. For all $j$, let $({\boldsymbol\omega}_{j}^\star,\mathbf s_{j}^\star)$ denote a limit point of $\{({\boldsymbol\omega}^{(t)},\mathbf{s}^{(t)})\}$ generated by Algorithm~\ref{alg:con-sample} with $c=c_j$.
\begin{Thm}[Convergence of Algorithm~\ref{alg:con-sample}]\label{thm:con-sample}
	Suppose that $f_{m}$, $m=0,\dots,M$ satisfy Assumption~\ref{asump:f}, $\bar{f}_{0}$ satisfies Assumption~\ref{asump:fbar}, $\bar{f}_{m}$ satisfies $\bar{f}_{m}(\boldsymbol\omega,\boldsymbol\omega,\mathbf{x})=f_{m}(\boldsymbol\omega,\mathbf{x})$ and Assumption~\ref{asump:fbar} for all $m=1,\dots,M$, the constraint set of Problem~\ref{Prob:con-sample} is compact, and the sequence $\{c_j\}$ satisfies $0<c_j<c_{j+1}$ and $\lim_{j\to\infty}c_j=\infty$. Then, the following statements hold.
	i) For all $j$, if $\mathbf s_{j}^\star=\mathbf0$, then ${\boldsymbol\omega}_{j}^\star$ is a KKT point of Problem~\ref{Prob:con-sample} almost surely;
	ii) A limit point of $\{({\boldsymbol\omega}_{j}^\star,\mathbf s_{j}^\star)\}$, denoted by $\{({\boldsymbol\omega}_{\infty}^\star,\mathbf s_{\infty}^\star)\}$, satisfies that $\mathbf s_{\infty}^\star=\mathbf{0}$, and ${\boldsymbol\omega}_{\infty}^\star$ is a KKT point of Problem~\ref{Prob:con-sample} almost surely.
\end{Thm}
\begin{IEEEproof}
	It follows from~\cite[Lemma1]{Lemma} that $\lim_{t\to\infty}\vert\bar{F}_{m}^t(\boldsymbol\omega^t)-
	F_{m}(\boldsymbol\omega^t)\Vert=0$ and $\lim_{t\to\infty} \Vert\nabla\bar{F}_{m}^t(\boldsymbol\omega^t)$  $-\nabla F_{m}(\boldsymbol\omega^t)\Vert=0$. Then, we can show the first statement by generalizing the analysis in \cite[Theorem~1]{Liu} and \cite[Theorem~2]{Ye}. Moreover, we can show the second statement by generalizing the proof of~\cite[Proposition~4.4.1]{bertsekas1998nonlinear}.
\end{IEEEproof}
In practice, we can choose a sequence $\{c_j\}$ which satisfies that $0<c_j<c_{j+1}$, $\lim_{j\to\infty}c_j=\infty$ and $c_1$ is large, and repeat Algorithm~\ref{alg:con-sample} with $c=c_j$ until $\Vert\mathbf s_{j}^\star\Vert$ is sufficiently small.

\subsection{Security Analysis}
We establish the security of Algorithm~\ref{alg:con-sample}. If for all $i\in\mathcal{I}$ and any mini-batch $\mathcal{N}'_i\subseteq\mathcal{N}_i$ with batch size $B\leq N_i$, the system of equations w.r.t. $\mathbf{z}\in\mathbb{R}^{BK}$, i.e., $\mathbf q_{m}\left(\boldsymbol\omega',\mathbf{z}\right)=\mathbf q_{m}\left(\boldsymbol\omega',(\mathbf{x}_n)_{n\in\mathcal{N}'_i}\right)$, $m=0,\dots,M$ has an infinite (or a sufficiently large) number of solutions, then raw data $\mathbf{x}_n$, $n\in\mathcal{N}^{(t)}_i$ cannot be extracted from $\mathbf q_{m}\left({\boldsymbol\omega}^{(t)},(\mathbf{x}_n)_{n\in\mathcal{N}^{(t)}_i}\right)$, $m=0,\dots,M$ in Step 4 of Algorithm~\ref{alg:con-sample}, and hence Algorithm~\ref{alg:con-sample} can preserve data privacy.
Otherwise, extra privacy mechanisms need to be explored. Note that federated learning with constrained optimization has not been studied so far, let alone the privacy mechanisms.

\subsection{Algorithm Example}
We provide an example of $\bar{f}_{m}$, $m=0,\dots,M$ with $\bar{f}_{0}$ satisfying Assumption~\ref{asump:fbar} and $\bar{f}_{m}$ satisfying $\bar{f}_{m}(\boldsymbol\omega,\boldsymbol\omega,\mathbf{x})=f_{m}(\boldsymbol\omega,\mathbf{x})$ and Assumption~\ref{asump:fbar} for all $m=1,\dots,M$.
Specifically, we can choose $\bar{f}_{0}$ given by \eqref{eqn:fs0bar} and choose $\bar{f}_{m}$, $m=1,\dots,M$ as follows:
\begin{align}
	\bar{f}_{m}(\boldsymbol\omega,{\boldsymbol\omega}^{(t)}\!,\mathbf{x}_n)\!=&f_{m}({\boldsymbol\omega}^{(t)}\!,\mathbf{x}_n)\!+\!\left(\nabla f_{m}({\boldsymbol\omega}^{(t)}\!,\mathbf{x}_n)\!\right)^T\!\!\left(\!\boldsymbol\omega\!-\!{\boldsymbol\omega}^{(t)}\!\right)\nonumber\\
	&+\tau\Vert{\boldsymbol\omega-{\boldsymbol\omega}^{(t)}}\Vert_2^2,\quad  m=1,\dots,M, \label{eqn:fsmbar}
\end{align}
where $\tau>0$ can be any constant.
Obviously, $\bar{f}_{0}$ given by \eqref{eqn:fs0bar} satisfies Assumption~\ref{asump:fbar}, and $\bar{f}_{m}$ given by~\eqref{eqn:fsmbar} satisfies $\bar{f}_{m}(\boldsymbol\omega,\boldsymbol\omega,\mathbf{x})=f_{m}(\boldsymbol\omega,\mathbf{x})$ and Assumption~\ref{asump:fbar} for all $m=1,\dots,M$.
Note that Problem~\ref{Prob:con-sample-ap} with $\bar{f}_{0}$ given by~\eqref{eqn:fs0bar} and $\bar{f}_{m}$, $m=1,\dots,M$ given by~\eqref{eqn:fsmbar} is a convex quadratically constrained quadratic programming, and can be solved using an interior point method.

\section{Application Examples}\label{sec:application}
In this section, we customize the proposed algorithmic frameworks to some applications and provide detailed solutions for the specific problems.
Define $\mathcal{K}\triangleq\{1,\dots,K\}$, $\mathcal{J}\triangleq\{1,\dots,J\}$ and $\mathcal{L}\triangleq\{1,\dots,L\}$.
Consider an $L$-class classification problem with a dataset of $N$ samples $(\mathbf{x}_n, \mathbf y_n)_{n\in\mathcal N}$, where $\mathbf{x}_n\triangleq(x_{n,k})_{k\in\mathcal{K}}$ and $\mathbf y_n\triangleq(y_{n,l})_{l\in\mathcal{L}}$ with $x_{n,k}\in\mathbb{R}$, and $y_{n,l}\in\{0,1\}$. Consider a three-layer neural network, including an input layer composed of $K$ cells, a hidden layer composed of $J$ cells, and an output layer composed of $L$ cells. 
We use the swish activation function $S(z)={z}/{(1+\exp(-z))}$~\cite{swish} for the hidden layer and the softmax activation function for the output layer.
We consider the cross entropy loss function. Thus, the resulting cost functions for sample-based and feature-based federated learning are given by:
\begin{align}
	F(\boldsymbol\omega)\triangleq
	-\frac{1}{N}\sum\limits_{n\in\mathcal{N}}\sum\limits_{l\in\mathcal{L}} y_{n,l}\log\left(Q_l(\boldsymbol\omega,\mathbf{x}_n)\right),
	\label{eqn:Fcost}
\end{align}
with $\boldsymbol\omega\triangleq({\omega}_{1,j,k},{\omega}_{2,l,j})_{k\in\mathcal{K},j\in\mathcal{J},l\in\mathcal{L}}$ and
\begin{align}
	&Q_l(\boldsymbol\omega,\mathbf{x}_n)\triangleq\frac{\exp(\sum_{j\in\mathcal{J}}{\omega}_{2,l,j} S(\sum_{k\in\mathcal{K}}{\omega}_{1,j,k}x_{n,k}))}{\sum_{h=1}^L \exp(\sum_{j\in\mathcal{J}}{\omega}_{2,h,j} S(\sum_{k\in\mathcal{K}}{\omega}_{1,j,k}x_{n,k}))},\nonumber\\
	&\hspace{6cm} l\in\mathcal{L}.\label{eqn:Ql}
\end{align}

\subsection{Unconstrained Federated Optimization}\label{sec:classification}
One unconstrained federated optimization formulation for the $L$-class classification problem is to minimize the weighted sum of the cost function $F(\boldsymbol\omega)$ in~\eqref{eqn:Fcost} together with the $\ell_2$-norm regularization term $\Vert\boldsymbol\omega\Vert^2_2$:
\begin{align}
	\min_{\boldsymbol\omega}\quad&F_{0}(\boldsymbol\omega)\triangleq F(\boldsymbol\omega)+\lambda\Vert\boldsymbol\omega\Vert^2_2\label{prob:class-uncon}
\end{align}
where $\lambda>0$ is the regularization parameter that trades off the cost and model sparsity.
We can apply Algorithm~\ref{alg:uncon-sample} with $\bar{f}_{0}(\boldsymbol\omega,{\boldsymbol\omega}^{(t)},\mathbf{x}_n)$ given by \eqref{eqn:fs0bar} to solve the problem in~\eqref{prob:class-uncon}. Theorem~\ref{thm:uncon-sample} guarantees the convergence of Algorithm~\ref{alg:uncon-sample}, as Assumption~\ref{asump:f} and Assumption~\ref{asump:fbar} are satisfied.
Specifically, the server solves the following convex approximate problem:
\begin{align}
	\min_{\boldsymbol\omega}\quad&\bar F_{0}^{(t)}(\boldsymbol\omega)=\bar F^{(t)}(\boldsymbol\omega)+2\lambda(\boldsymbol\beta^{(t)})^T\boldsymbol\omega\label{prob:class-uncon-ap}
\end{align}
where
$\bar F^{(t)}(\boldsymbol\omega)$ is given by 
\begin{align}
	\bar F^{(t)}(\boldsymbol\omega)=\sum_{j\in\mathcal{J}}\sum_{k\in\mathcal{K}}{B_{j,k}^{(t)}}{\omega}_{1,j,k}
	+\sum_{l\in\mathcal{L}}\sum_{j\in\mathcal{J}} {C_{l,j}^{(t)}}{\omega}_{2,l,j}
	+\tau\Vert{\boldsymbol\omega}\Vert_2^2,\label{eqn:fbar-app}
\end{align}
and $\boldsymbol\beta^{(t)}\in\mathbb{R}^d$, ${B_{j,k}^{(t)}}$ and ${C_{l,j}^{(t)}}$ are updated according to:
\begin{align}
	&\boldsymbol\beta^{(t)}=(1-\rho^{(t)})\boldsymbol\beta^{(t-1)}+\rho^{(t)}{\boldsymbol\omega}^{(t)},\nonumber\\
	&{B_{j,k}^{(t)}}\!=\!(1-\rho^{(t)}){B_{j,k}^{(t-1)}}\!+\rho^{(t)}\!\left(\bar{B}_{j,k}^{(t)}\!-2\tau\omega^{(t)}_{1,j,k}\right),\label{eqn:B}\\
	&{C_{l,j}^{(t)}}=(1-\rho^{(t)}){C_{l,j}^{(t-1)}}+\rho^{(t)}\left(\bar{C}_{l,j}^{(t)}-2\tau\omega^{(t)}_{2,l,j}\right),\label{eqn:C}
\end{align}
respectively, with ${\boldsymbol{\beta}}^{(0)}=\mathbf 0$ and ${B_{j,k}^{(0)}}={C_{l,j}^{(0)}}=0$.
Here, $\bar{B}_{j,k}^{(t)}$ and $\bar{C}_{l,j}^{(t)}$ are given by:
\begin{align}
	\bar{B}_{j,k}^{(t)}\!=&
	\sum_{i\in\mathcal{I}}\frac{N_i}{BN}\sum_{n\in\mathcal N_i^{(t)}}\sum_{l\in\mathcal{L}}\left(Q_l({\boldsymbol\omega}^{(t)},\mathbf{x}_n)-y_{n,l}\right)\nonumber\\
	&\times S'\left(\sum_{k'=1}^K\omega^{(t)}_{1,j,k'}x_{n,k'}\right)\omega^{(t)}_{2,l,j}x_{n,k},\nonumber\\
	\bar{C}_{l,j}^{(t)}\!=&
	\sum_{i\in\mathcal{I}}\!\frac{N_i}{BN}\!\!\!\!\sum_{n\in\mathcal N_i^{(t)}}\!\!\!\left(\!Q_l({\boldsymbol\omega}^{(t)}\!,\mathbf{x}_n)\!-\!y_{n,l}\right)\!S\!\left(\sum_{k'=1}^K\!\omega^{(t)}_{1,j,k'}x_{n,k'}\!\!\right)\!,\nonumber
\end{align}

By the first-order optimality condition, the closed-form solution of the problem in~\eqref{prob:class-uncon-ap} is given by:
\begin{align}
	&\bar{\omega}_{1,j,k}^{(t)}=-\frac{1}{2\tau}\left({B_{j,k}^{(t)}}+2\lambda{\beta}_{1,j,k}^{(t)}\right),\  j\in\mathcal{J},\ k\in\mathcal{K},\label{eqn:omega1-uncon}\\
	&\bar{\omega}_{2,l,j}^{(t)}=-\frac{1}{2\tau}\left({C_{l,j}^{(t)}}+2\lambda{\beta}_{2,l,j}^{(t)}\right),\ l\in\mathcal{L},\ j\in\mathcal{J}.\label{eqn:omega2-uncon}
\end{align}
Thus, in Step 5 in Algorithm~\ref{alg:uncon-sample}, the server only needs to compute $\boldsymbol\omega$ according to~\eqref{eqn:omega1-uncon} and~\eqref{eqn:omega2-uncon}, respectively.

\subsection{Constrained Federated Optimization}
One constrained federated optimization formulation for the $L$-class classification problem is to minimize the $\ell_2$-norm of the network parameters $\Vert\boldsymbol\omega\Vert^2_2$ under a constraint on the cost function $F(\boldsymbol\omega)$ in~\eqref{eqn:Fcost}:
\begin{align}
	\min_{\boldsymbol\omega}\quad& F_{0}(\boldsymbol\omega)\triangleq\Vert\boldsymbol\omega\Vert^2_2\label{prob:class-con}\\
	\text{s.t.}\quad &F_{1}(\boldsymbol\omega)\triangleq F(\boldsymbol\omega)-U\leq0,\nonumber
\end{align}
where $U$ represents the limit on the cost.
We can apply Algorithm~\ref{alg:con-sample} with $\bar{f}_{0}(\boldsymbol\omega,{\boldsymbol\omega}^{(t)},\mathbf{x}_n)$ given by \eqref{eqn:fs0bar} and $\bar{f}_{m}(\boldsymbol\omega,{\boldsymbol\omega}^{(t)},\mathbf{x}_n)$ given by \eqref{eqn:fsmbar} to solve the problem in~\eqref{prob:class-con}. The convergence of Algorithm~\ref{alg:con-sample} is guaranteed by Theorem~\ref{thm:con-sample}, as Assumption~\ref{asump:f} and Assumption~\ref{asump:fbar} are satisfied.
Specifically, the server solves the following convex approximate problem:
\begin{align}
	\min_{\boldsymbol\omega,s}\quad&\Vert\boldsymbol\omega\Vert^2_2+c s\label{prob:class-con-ap}\\
	\text{s.t.}\quad &\bar F^{(t)}(\boldsymbol\omega)+A^{(t)}-U\leq s,\nonumber\\
	&s\geq0, \nonumber
\end{align}
where $\bar F^{(t)}(\boldsymbol\omega)$ is given by~\eqref{eqn:fbar-app} with ${B_{j,k}^{(t)}}$, ${C_{l,j}^{(t)}}$ and $A^{(t)}$ updated according to~\eqref{eqn:B}, \eqref{eqn:C} and
\begin{align}
	&A^{(t)}=(1-\rho^{(t)})A^{(t-1)}+\nonumber\\
	&\rho^{(t)}\bigg(\bar{A}^{(t)}-\sum_{j\in\mathcal{J}}\sum_{k\in\mathcal{K}} \bar{B}_{j,k}^{(t)}\omega^{(t)}_{1,j,k}
	-\sum_{l\in\mathcal{L}}\sum_{j\in\mathcal{J}} \bar{C}_{l,j}^{(t)}\omega^{(t)}_{2,l,j}\bigg), \label{eqn:A}
\end{align}
respectively, with $A^{(0)}=0$ and $\bar{A}^{(t)}$ given by:
\begin{align}
	&\bar{A}^{(t)}\!=\!
	\sum_{i\in\mathcal{I}}\frac{N_i}{BN}\!\!\sum_{n\in\mathcal N_i^{(t)}}\sum_{l\in\mathcal{L}} y_{n,l}\log\left(Q_l({\boldsymbol\omega}^{(t)},\mathbf{x}_n)\right)\!+\!\tau\Vert{{\boldsymbol\omega}^{(t)}}\Vert_2^2,\nonumber
\end{align}

By the KKT conditions, the closed-form solutions of the problem in~\eqref{prob:class-con-ap} is given as follows.
\begin{Lem}[Optimal Solution of Problem in~\eqref{prob:class-con-ap}]\label{lem:closedform}
	\begin{align}
		&\bar{\omega}_{1,j,k}^{(t)}=\frac{-\nu B_{j,k}^{(t)}}{2(1+\nu\tau)},\quad j\in\mathcal{J},\ k\in\mathcal{K},\label{eqn:omega1-con}\\
		&\bar{\omega}_{2,l,j}^{(t)}=\frac{-\nu C_{l,j}^{(t)}}{2(1+\nu\tau)},\quad l\in\mathcal{L},\ j\in\mathcal{J},\label{eqn:omega2-con}
	\end{align}
	where
	\begin{align}
		&\nu=
		\begin{cases}
			\left[\frac{1}{\tau}\left(\sqrt{\frac{b}{b+4\tau(U-A^{(t)})}}\!-\!1\right)\right]_0^c,&b+4\tau(U-A^{(t)})>0\\
			c,&b+4\tau(U-A^{(t)})\leq0,
		\end{cases}\nonumber\\
		&b=\sum_{j\in\mathcal{J}}\sum_{k\in\mathcal{K}} (B_{j,k}^{(t)})^2+\sum_{l\in\mathcal{L}}\sum_{j\in\mathcal{J}} (C_{l,j}^{(t)})^2.\label{eqn:b}
	\end{align}
	Here, $[x]^c_0\triangleq\min\left\{\max\{x,0\},c\right\}$.
\end{Lem}
Thus, in Step 5 of Algorithm~\ref{alg:con-sample}, the server only needs to compute $\boldsymbol\omega$ according to~\eqref{eqn:omega1-con} and~\eqref{eqn:omega2-con}.

\section{Numerical Results}\label{sec:simu}
In this section, we show the performance of Algorithm~\ref{alg:uncon-sample}, Algorithm~\ref{alg:con-sample} and the SGD-based algorithms~\cite{mcmahan2017communication,yang2019scheduling,yu2019parallel} in the application examples in Sections~\ref{sec:application} using numerical experiments.
We carry our experiments on Mnist data set.
For the training model, we choose, $N=60000$, $I=10$, $K=784$, $J=128$, $L=10$.
For Algorithm~\ref{alg:uncon-sample} and Algorithm~\ref{alg:con-sample}, we choose $T=100$, $\tau=0.1$, $c=10^5$, 
$\rho^t=a_1/t^{\alpha}$ and $\gamma^t=a_2/t^{\alpha+0.05}$ with $a_1=0.4,0.6,0.9$, $a_2=0.4,0.9,0.9$, $\alpha=0.4,0.3,0.3$ for batch sizes $B=1,10,100$, respectively.
For the SGD-based algorithms~\cite{mcmahan2017communication,yang2019scheduling,yu2019parallel}, let $E$ denote the number of local SGD updates, and the learning rate is set as $r=\bar a/t^{\bar\alpha}$, where $\bar a$ and $\bar\alpha$ are selected using grid search method.
Note that all the results are given by the average over 100 runs.

\begin{figure}[h]
	\begin{center}
		\subfigure[\scriptsize{
			Training cost $F(\boldsymbol\omega^t)$ vs. iteration $t$ by Algorithm~\ref{alg:uncon-sample} with $\lambda=10^{-5}$.}\label{fig:CostSam}]
		{\resizebox{4.2cm}{!}{\includegraphics{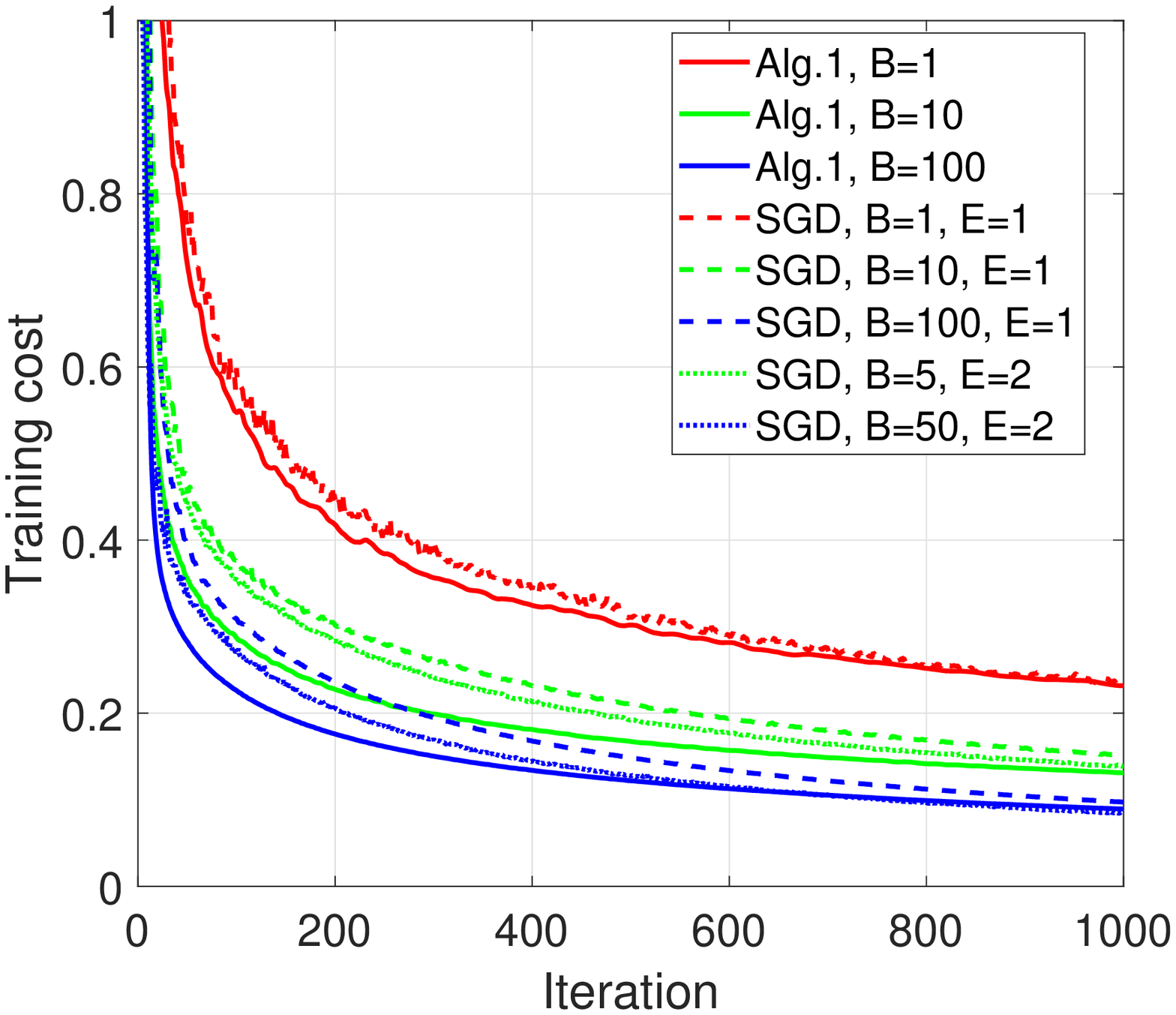}}}\quad
		\subfigure[\scriptsize{
			Training cost $F(\boldsymbol\omega^t)$ vs. iteration $t$ by Algorithm~\ref{alg:con-sample} with $U=0.13$.}\label{fig:CostSamCon}]
		{\resizebox{4.2cm}{!}{\includegraphics{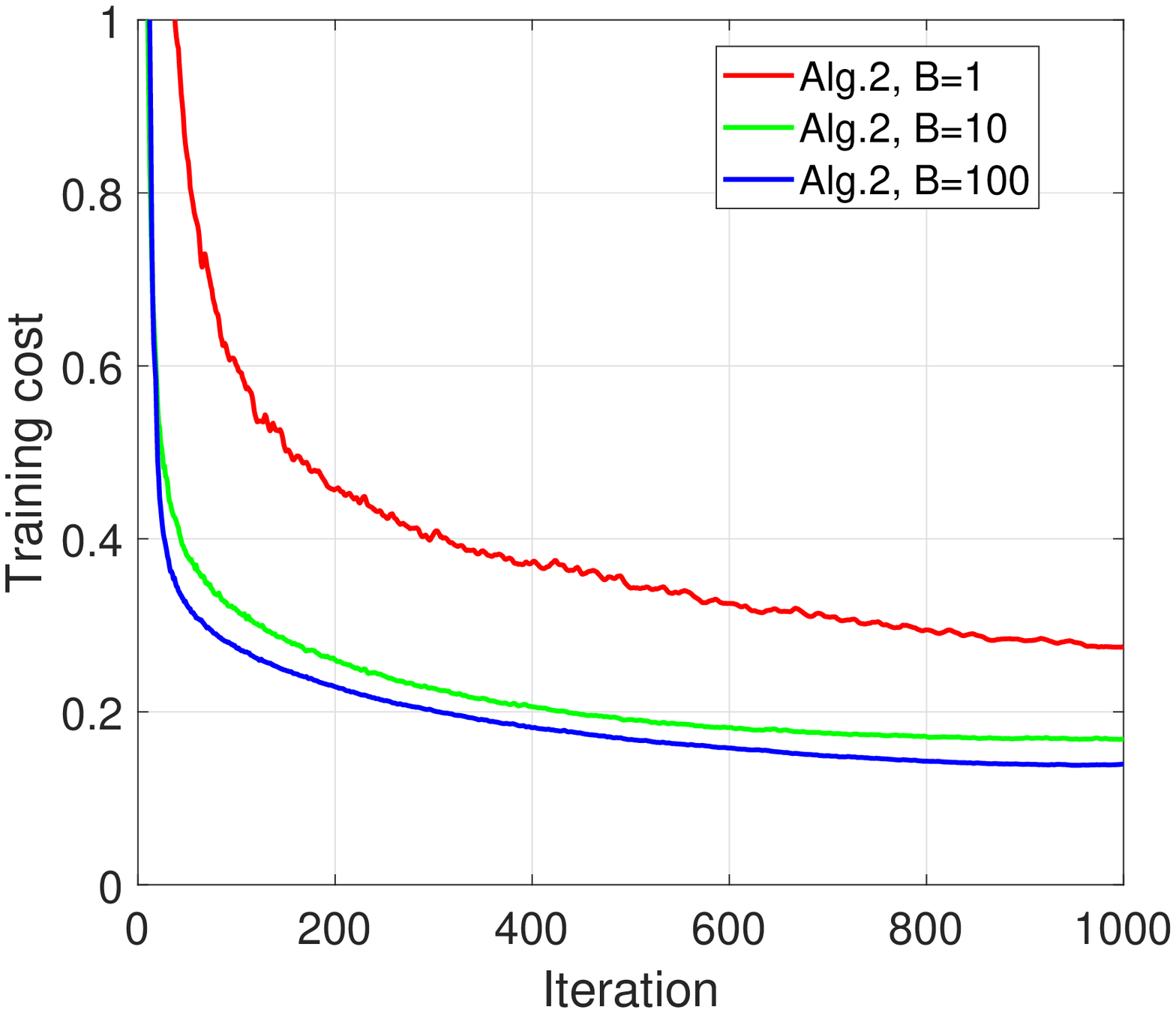}}}\\
	\end{center}
\vspace{-4mm}
	\caption{\small{Training cost versus iteration index.}}
	\label{fig:Cost}
\end{figure}
\vspace{-4mm}

\begin{figure}[h]
	\begin{center}
		\subfigure[\scriptsize{
			Test accuracy at $\boldsymbol\omega^t$ vs. iteration $t$ by Algorithm~\ref{alg:uncon-sample} with $\lambda=10^{-5}$.}\label{fig:AccuSam}]
		{\resizebox{4.2cm}{!}{\includegraphics{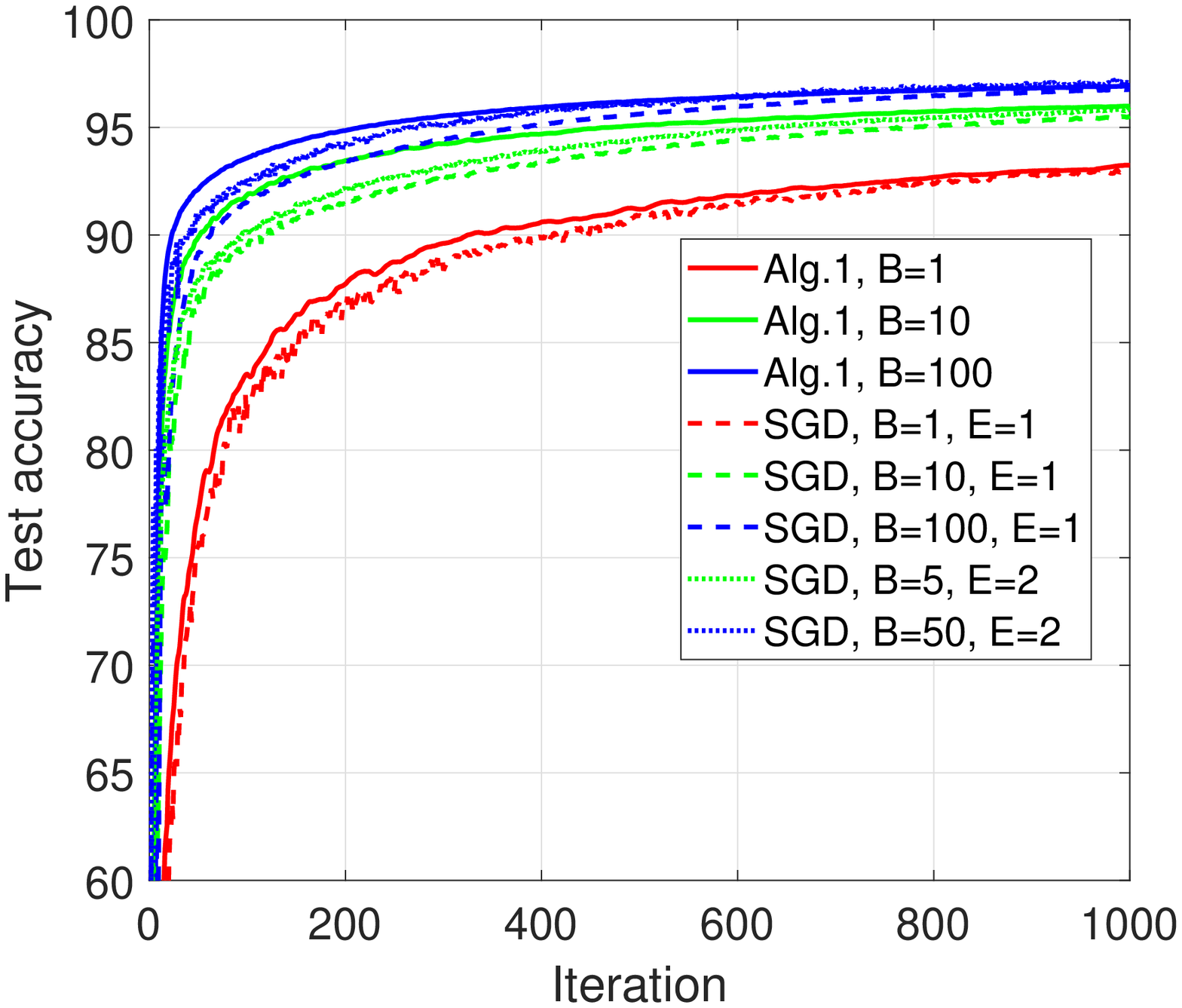}}}\quad
		\subfigure[\scriptsize{
			Test accuracy at $\boldsymbol\omega^t$ vs. iteration $t$ by Algorithm~\ref{alg:con-sample} with $U=0.13$.}\label{fig:AccuSamcon}]
		{\resizebox{4.2cm}{!}{\includegraphics{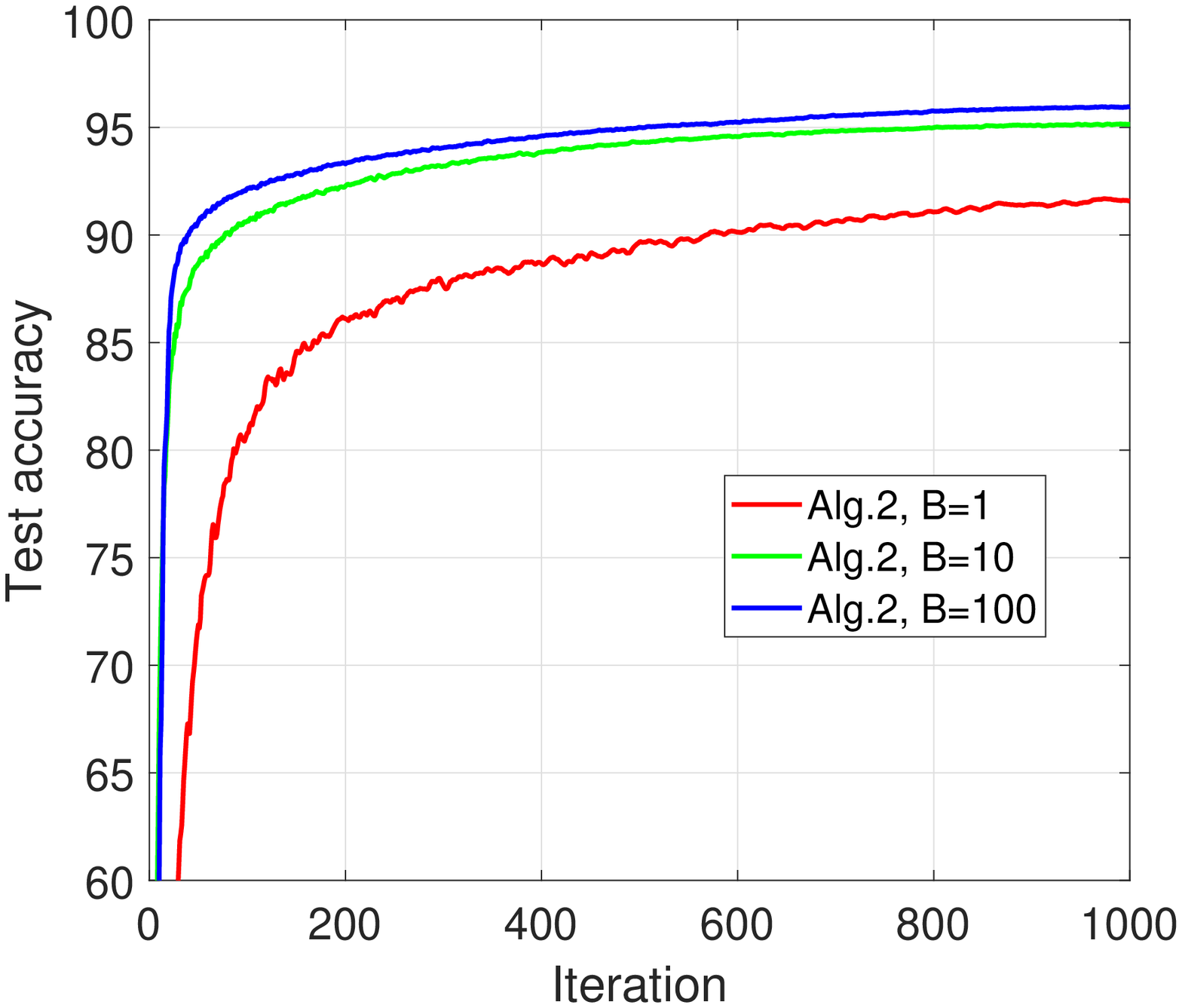}}}\\
	\end{center}
\vspace{-4mm}
	\caption{\small{Test accuracy versus iteration index.}}
	\label{fig:Accu}
\end{figure}
\vspace{-4mm}

\begin{figure}[h]
	\begin{center}
		\subfigure[\scriptsize{
			$\ell_2$-norm $\Vert\boldsymbol\omega\Vert^2_2$ vs. training cost obtained by Algorithm~\ref{alg:uncon-sample}.}\label{fig:Tradeoff1}]
		{\resizebox{4.2cm}{!}{\includegraphics{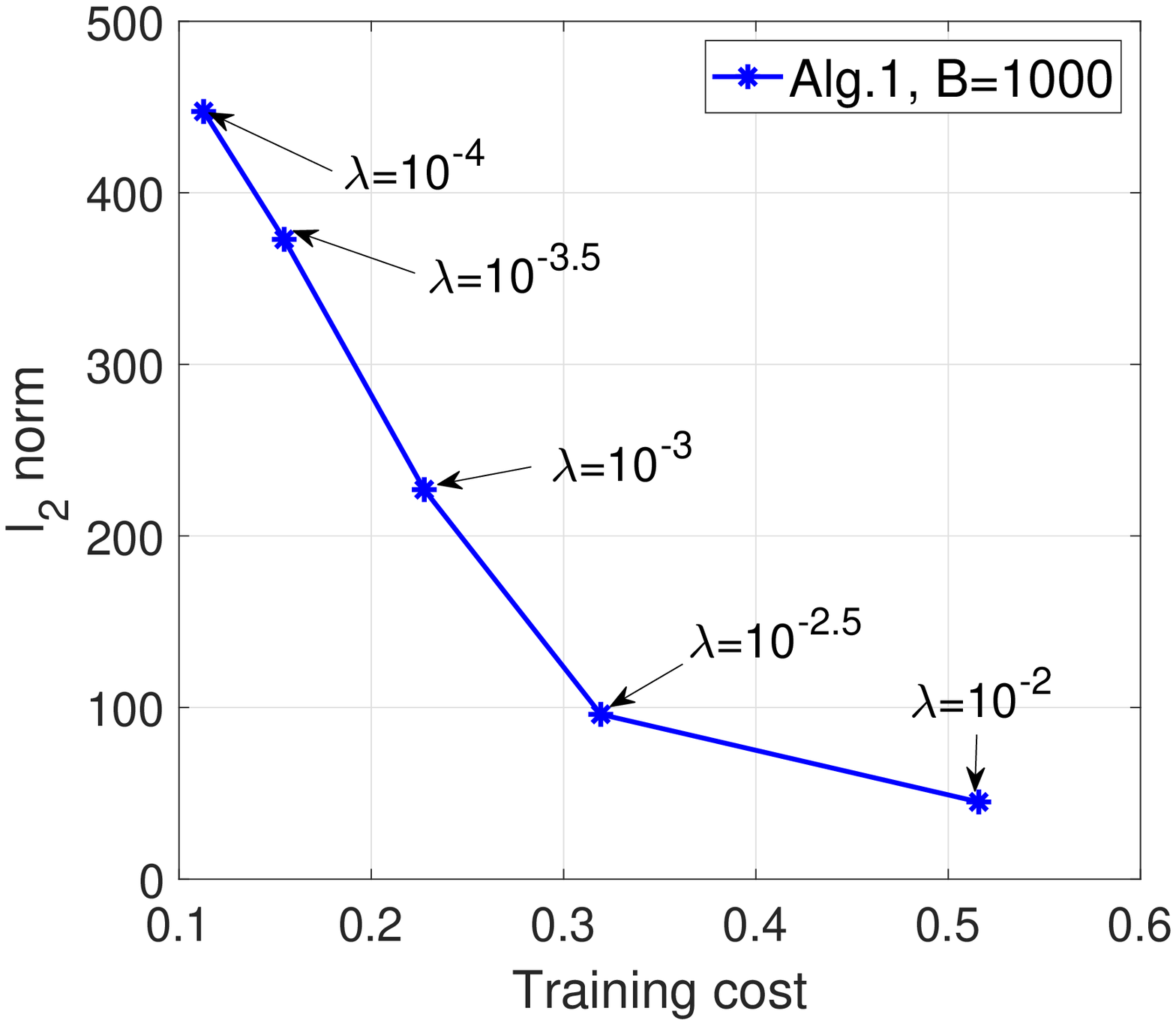}}}\quad
		\subfigure[\scriptsize{
			$\ell_2$-norm $\Vert\boldsymbol\omega\Vert^2_2$ vs. training cost obtained by Algorithm~\ref{alg:con-sample}.}\label{fig:Tradeoff2}]
		{\resizebox{4.2cm}{!}{\includegraphics{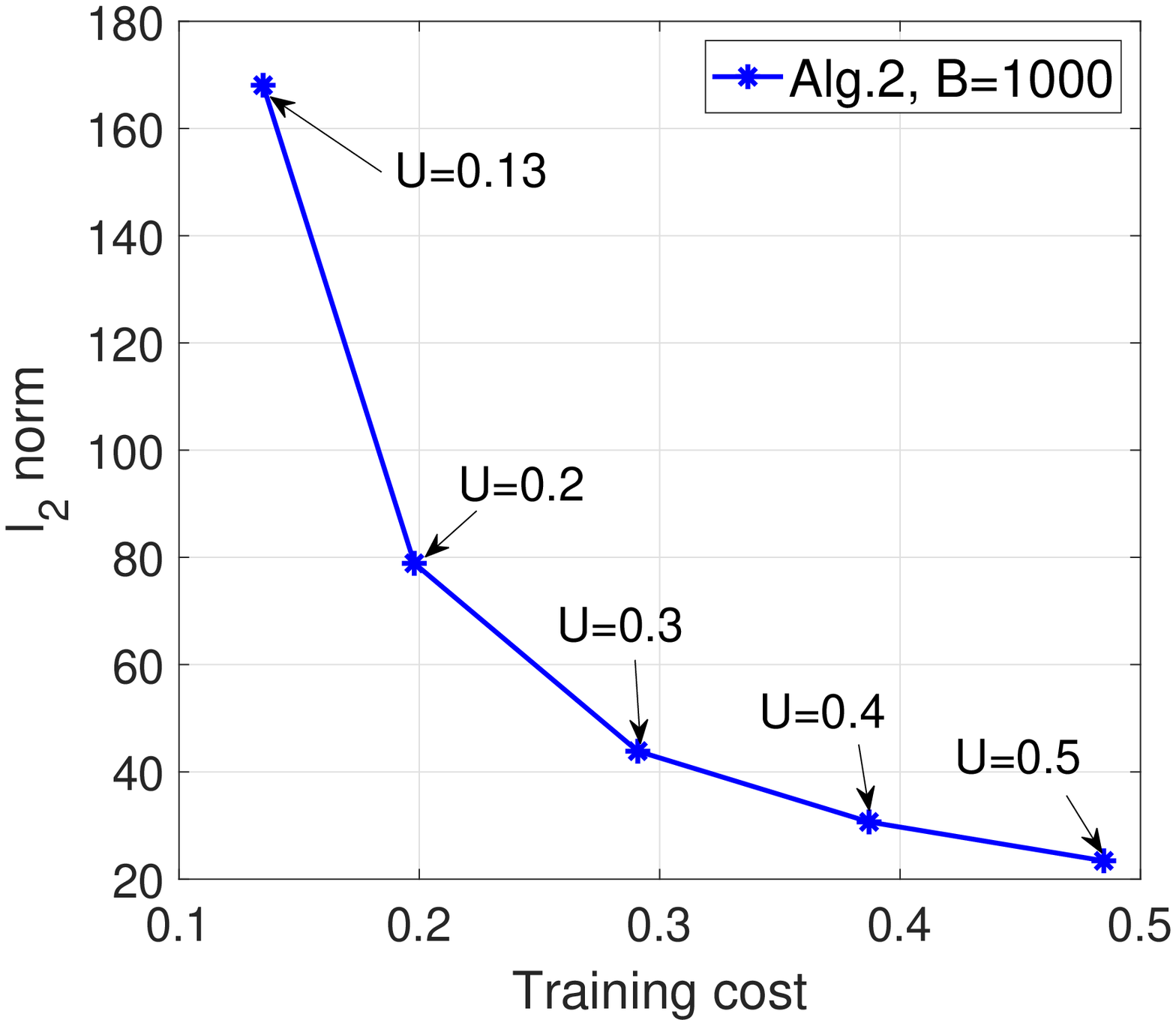}}}
	\end{center}
\vspace{-4mm}
	\caption{\small{Model sparsity versus training cost.}}
	\label{fig:Spar}
\end{figure}

Fig.~\ref{fig:Cost} and Fig.~\ref{fig:Accu} illustrate the training cost and test accuracy versus the iteration index. From Fig.~\ref{fig:Cost} and Fig.~\ref{fig:Accu}, we can see that the proposed algorithms with larger batch sizes converge faster. From Fig.~\ref{fig:CostSam} and Fig.~\ref{fig:AccuSam}, we can observe that for unconstrained federated optimization, Algorithm~\ref{alg:uncon-sample} converges faster than the SGD-based algorithm with $E=1$ at the same batch size. 
In addition, Algorithm~\ref{alg:uncon-sample} with $B=10(100)$ converges faster than the SGD-based algorithm with $B=5(50)$ and $E=2$, i.e., Algorithm~\ref{alg:uncon-sample} converges faster that the SGD-based algorithm when the two algorithms induce the same computation load for each client.
Fig.~\ref{fig:Tradeoff1} and Fig.~\ref{fig:Tradeoff2} show the tradeoff curve between the model sparsity and training cost of each proposed algorithm. From Fig.~\ref{fig:Tradeoff2}, we see that with constrained sample-based federated optimization, one can set an explicit constraint on the training cost to effectively control the test accuracy. Furthermore, by comparing Fig.~\ref{fig:Tradeoff1} and Fig.~\ref{fig:Tradeoff2}, we can see that Algorithm~\ref{alg:con-sample} can achieve a better tradeoff between the model sparsity and training cost than Algorithm~\ref{alg:uncon-sample}. The main reason is that the underlying constrained sample-based federated optimization has a convex objective function and the chance for Algorithm~\ref{alg:con-sample} to converge to an optimal point is higher.

\section{Conclusions}
In this paper, we proposed two privacy preserving algorithms for unconstrained and constrained sample-based federated optimization problems, respectively, using SSCA techniques. We also showed that each algorithm can converge to a KKT point of the corresponding problem. It is worth noting that SSCA has not been used for solving federated optimization, and federated optimization with nonconvex constraints has not been investigated. Numerical experiments showed that the proposed SSCA-based algorithm for unconstrained sample-based federated optimization converges faster than the existing SGD-based algorithms, and the proposed SSCA-based algorithm for constrained sample-based federated optimization can obtain a sparser model that satisfies an explicit constraint on the model cost.


\end{document}